\documentclass[letterpaper, 10 pt, conference]{ieeeconf}

\IEEEoverridecommandlockouts
\usepackage[english]{babel}
\usepackage{subfigure}
\usepackage{bm}
\usepackage{subfig}
\usepackage{amsmath}
\usepackage{tensor}
\usepackage{amsfonts}
\usepackage{algorithm}
\usepackage{algorithmic}
\usepackage{url}
\usepackage{amssymb}
\usepackage{graphicx}
\usepackage{amsmath}
\usepackage{breqn}
\usepackage{siunitx}
\usepackage{afterpage}
\usepackage{soul}
\usepackage{multirow}
\usepackage{here} 
\newcolumntype{b}{X}
\newcolumntype{s}{>{\hsize=.3\hsize}X}
\newcolumntype{x}{>{\hsize=.5\hsize}X}
\usepackage{tikz}
\usetikzlibrary{shapes.geometric, arrows}
\usepackage{color, colortbl}
\definecolor{Gray}{gray}{0.9}
\usepackage[toc,sort=use]{glossaries}
\usepackage{array}
\newglossary[slg]{symbols}{sym}{sbl}{List of Symbols}
\newglossary[slg]{hidden}{sym}{sbl}{Hidden Symbols}
\makeglossaries
         
\makeatletter

\usepackage{cite}
\usepackage{wasysym}
\title{\LARGE \bf VIPose: Real-time Visual-Inertial 6D Object Pose Tracking}

\author{
    Rundong Ge and Giuseppe Loianno
\thanks{The authors are with the New York University, Tandon School of Engineering, Brooklyn, NY 11201, USA. {\tt\footnotesize email: \{rundong.ge, loiannog\}@nyu.edu}.}
\thanks{This work was supported by Qualcomm Research, the Technology Innovation Institute, Nokia, NYU Wireless, and the young researchers program "Rita Levi di Montalcini" 2017 grant PGR17W9W4N.}
}

\begin{document}

\maketitle
\thispagestyle{empty}
\pagestyle{empty}

\begin{abstract}
Estimating the 6D pose of objects is beneficial for robotics tasks such as transportation, autonomous navigation, manipulation as well as in scenarios beyond robotics like virtual and augmented reality. With respect to single image pose estimation, pose tracking takes into account the temporal information across multiple frames to overcome possible detection inconsistencies and to improve the pose estimation efficiency. In this work, we introduce a novel Deep Neural Network (DNN) called VIPose, that combines inertial and camera data to address the object pose tracking problem in real-time. The key contribution is the design of a novel DNN architecture which fuses visual and inertial features to predict the objects' relative 6D pose between consecutive image frames. The overall 6D pose is then estimated by consecutively combining relative poses. Our approach shows remarkable pose estimation results for heavily occluded objects that are well known to be very challenging to handle by existing state-of-the-art solutions. The effectiveness of the proposed approach is validated on a new dataset called VIYCB with RGB image, IMU data, and accurate 6D pose annotations created by employing an automated labeling technique. The approach presents accuracy performances comparable to state-of-the-art techniques, but with the additional benefit of being real-time.

\end{abstract}

\IEEEpeerreviewmaketitle

\section{Introduction}~\label{sec:introduction}
Tracking the 6D pose of objects from RGB image sequences, (i.e., estimating 3D translation and rotation of objects with respect to the camera in each frame), is an important task in various robotics applications. This can boost robots' navigation performances in various outdoor and indoor settings to solve complex tasks such as inspection, mapping, and search and rescue. It can also be used in object manipulation tasks for planning and grasping, aerial cinematography to track moving targets as well as in scenarios beyond robotics such as Virtual Reality (VR) and Augmented Reality (AR).%

Estimating the 6D pose of objects from a single RGB image has been extensively studied. Traditional methods match objects' keypoints across images and 3D object models~\cite{collet2011moped,hinterstoisser2011gradient,linemod}. Recently deep learning-based methods have significantly improved the accuracy and robustness of pose estimation \cite{ssd6d,tekin,zeng2017multi,pavlakos20176,posecnn,deepim,pvnet,cdpn}. Several single image pose estimation methods are real-time ($30$ Hz) whereas the accuracy is limited~\cite{tekin,cdpn}. Furthermore, these methods ignore the temporal and spatial information across consecutive image frames and strictly focus on single-view pose estimation. This may lead to inconsistent pose estimations across consecutive frames. 

\begin{figure}[t!]
    \centering
    \includegraphics[width=1.0\linewidth]{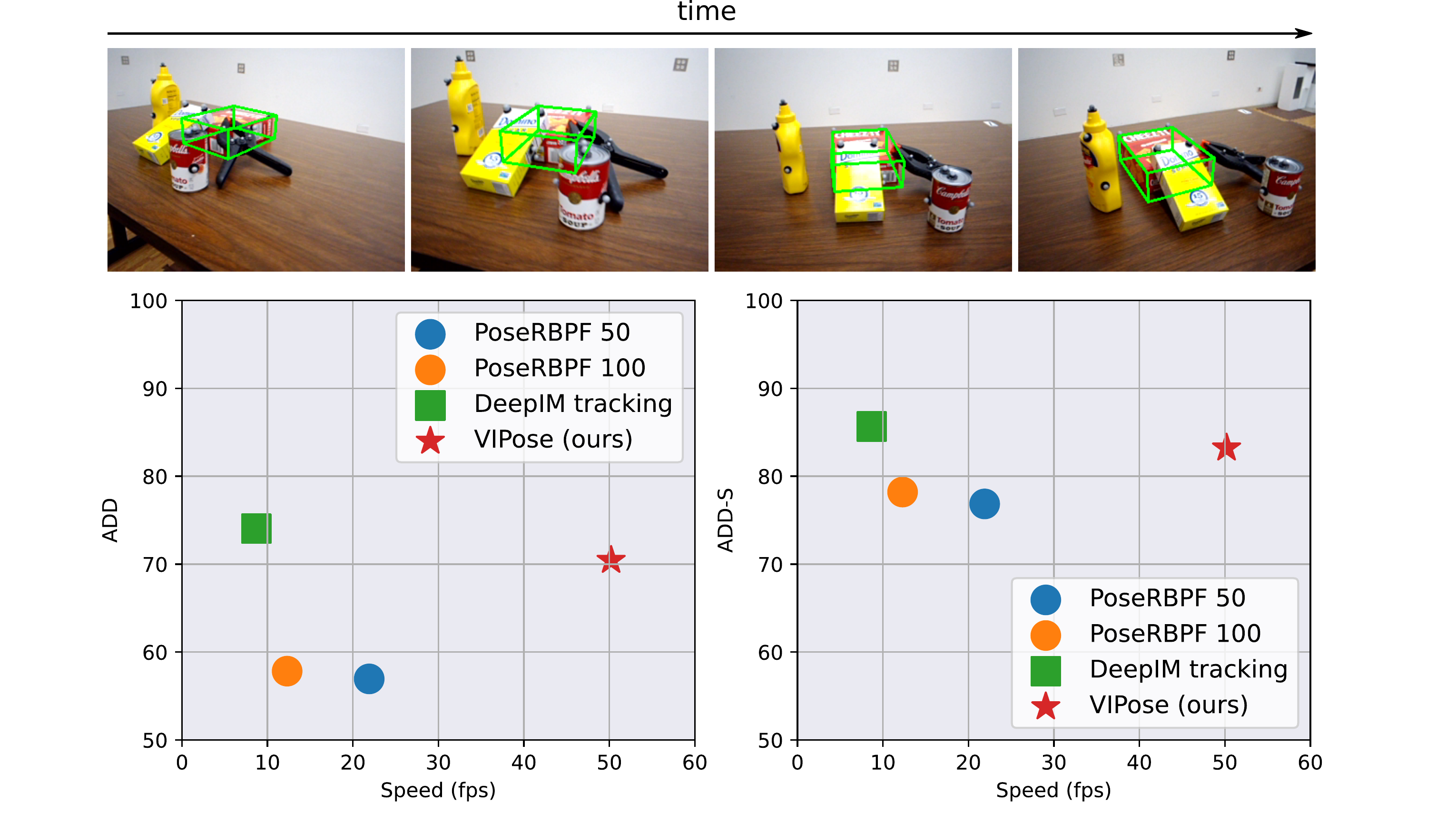}
    \caption{Visual-Inertial  6D  object  pose  tracking (top). Speed-accuracy benchmark on our VIYCB dataset with respect to state-of-the-art solutions (bottom).}
    \label{fig:fig1}
    \vspace{-10pt}
\end{figure}

Recent RGB-based 6D object pose tracking methods achieved good tracking performances~\cite{deepim,poserbpf}. However, these methods are not robust to severe occlusions, which enforce the re-initialization of the pose tracking process each time this event occurs. Re-initialization is usually performed by an accurate 6D pose estimation method, which is computationally expensive. Therefore, the speed of these approaches is quite limited and prevents their application in real-time scenarios.

To overcome the aforementioned drawbacks, we propose a novel DNN architecture, which combines Inertial Measurement Unit (IMU) data with camera images. The IMU has been widely used in the robotics field for localization and navigation. Recent research \cite{ronin,tlio} show that IMU data can be processed by a trained DNN to predict the velocity, which can be leveraged to resolve the camera localization problem. Therefore, the IMU sensor can provide useful information on the camera motion. In our case, we exploit this information to achieve accurate object pose tracking even under severe occlusions. In addition, the joint learning process on images and IMU data leads to an efficient representation that enables robust real-time pose tracking. 
\begin{figure*}[!t]
    \centering
    \includegraphics[width=0.9\linewidth]{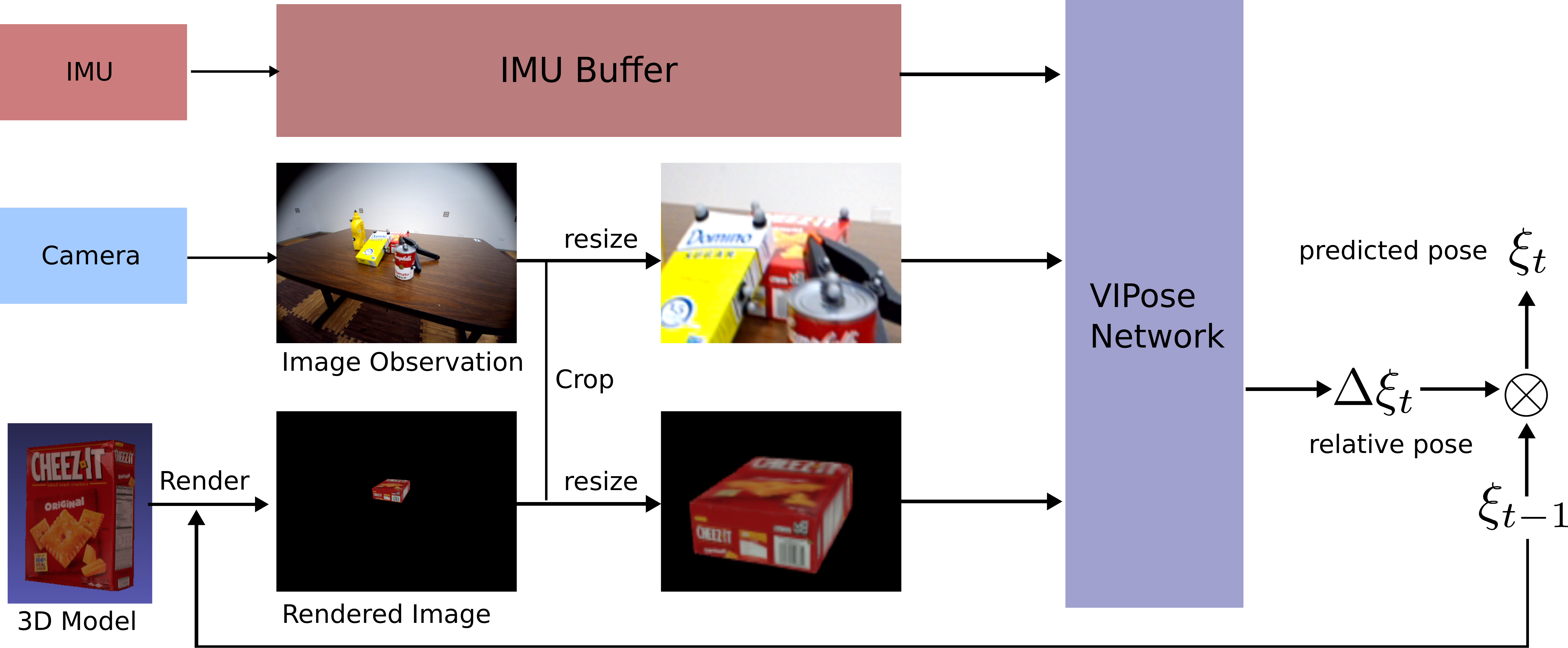}
    \caption{Pose Tracking Workflow. At time t, the camera provides the image observation $I_t$. The 3D model of the object and the pose estimation of last time $\bm{\xi}_{t-1}$ are used to render $I_{t-1}^R$. $I_t$ and $I_{t-1}^R$ are both cropped by an enlarged bounding box (illustrated in Fig.~\ref{fig:bbox}) and resized to half of original resolution. The IMU buffer provides the recent $K$ IMU measurements. The pre-processed image pair and the IMU sequence are fed into the VIPose network to predict the relative pose $\Delta \bm{\xi}_t$ which is combined with $\bm{\xi}_{t-1}$ to obtain the object pose estimation $\bm{\xi}_t$ at this frame.}
    \label{fig:pipeline}
    \vspace{-15pt}
\end{figure*}

This work presents multiple contributions. First, we propose the first object pose tracking method that fuses camera and IMU data to address the challenge of pose tracking of objects in complex scenes that include objects' occlusions. Second, our approach speeds up the pose tracking problem by leveraging IMU data and inferring the overall pose by propagating and combining relative pose estimation across consecutive frames. Finally, we experimentally validate our approach on collected dataset including RGB images and IMU data that is generally not available in existing datasets. To overcome the well-known difficult task of generating 6D pose annotations, an automated labeling technique is also introduced to collect enough training and testing data thus significantly reducing manually labeling effort.
Experimental results show that the proposed approach achieves a comparable result with respect to existing state-of-the-art solutions, while running significantly faster and achieving real-time performance (i.e $50.2$ Hz).

The paper is organized as follows. In Section~\ref{sec:related_works}, we review related work on 6D pose estimation and tracking, as well as learning techniques for localization based on IMU data. Section~\ref{sec:approach} introduces the proposed tracking pipeline and the VIPose network. Section~\ref{sec:experiments} presents extensive experimental results on our VIYCB dataset. Section~\ref{sec:conclusion} concludes the work and presents future directions.

\section{Related Work}~\label{sec:related_works}

\textbf{6D Pose Estimation:} Traditionally, the 6D pose of an object in a single image is estimated with template matching methods, where feature points in an image are matched with the corresponding 3D model feature points. The 6D object pose can then be estimated by solving 2D-3D correspondences of these local features using the PnP algorithm~\cite{collet2011moped}. Recently, deep learning-based methods have surged in popularity. SSD-6D~\cite{ssd6d} extends the SSD object detector to 6D pose estimation by adding a viewpoint classification branch.~\cite{tekin} adopts the YOLO network to detect 3D bounding box corners of objects in images and recover the 6D pose by again solving a PnP problem. PoseCNN~\cite{posecnn} proposes an end-to-end framework to perform semantic segmentation and predict 3D rotation and translation. Conversely, DeepIM~\cite{deepim} adopts the FlowNetSimple network~\cite{flownet} to refine the estimation of single image 6D pose detection methods by iteratively predicting the relative pose from the rendered previous pose to the current image observation until the iterative pose increment is small enough. Generally, adopting 6D pose estimation methods to estimate 6D pose for each frame in videos ignores temporal information which can be useful to improve the speed and accuracy of the pose estimation process without performing estimation at each frame.

\textbf{6D Pose Tracking:} Our work is closely related to recent advances in RGB-based 6D object pose tracking. PWP3D~\cite{pwp3d} proposes to track the object pose by optimizing the projected contour of the 3D model of the object. \cite{tjaden2016real} further improves~\cite{pwp3d} with a novel optimization scheme and GPU parallelization. Then \cite{tjaden2017real} proposes to improve pose tracking using a temporally local color histogram. Recently, \cite{manhardt2018deep} adopts a deep neural network to predict the pose difference between consecutive frames. PoseRBPF~\cite{poserbpf} proposes a Rao-Blackwellized particle filter approach for 6D pose tracking, which decouples 3D rotation and 3D translation and takes the uncertainty and object symmetry into account, achieving state-of-the-art performances on the YCB Video Dataset. DeepIM~\cite{deepim} can be extended to perform pose tracking by predicting the relative transformation between two consecutive frames. However, these methods are still not robust to severe occlusions and re-initialization is required when the tracking is lost. \cite{se3tracknet} proposes a novel network for 6D pose tracking on RGB-D data. It disentangles feature encoding to reduce the domain gap in simulation and reality and only uses synthetic data for training. This method performs well on RGB-D data whereas in our work we focus on the more challenging scenarios where only RGB images are available without the depth information.

\textbf{Data-driven IMU Localization:} Traditionally, IMU data can be processed by some filter-based approach for localization. In recent years, deep learning methods have been applied to regress localization information directly from IMU data. VINet~\cite{vinet} proposes the first end-to-end system for visual-inertial odometry, which adopts an LSTM~\cite{lstm} network to process IMU input across image frames and a FlowNet network to process image pairs. The visual-inertial fusion is obtained by concatenating the IMU and visual feature vectors. IONet~\cite{ionet} proposes an LSTM network to predict relative displacement in 2D and concatenates the predictions to infer the position relying only on the IMU input. RoNIN~\cite{ronin} rotates the IMU data into a gravity-aligned frame and predicts the average 2D velocity over a sequence of IMU data employing a 1D ResNet network. TLIO~\cite{tlio} adopts 1D ResNet network as well, but to regress 3D displacements and corresponding uncertainties with a tightly-coupled Extended Kalman Filter framework for state estimation. Our work closely relates to the aforementioned works on IMU data using 1D ResNet, but we focus on object pose and tracking problems and not on camera localization.

\section{Approach}~\label{sec:approach}
The goal of the pose tracking problem is to estimate the 6D pose of the object with respect to the camera frame $\bm{\xi}_t\in SE(3)$ at time $t,~t\in\{1,2,\cdots,N\}$ given
\begin{itemize}
    \item A 3D CAD model of the object
    \item The initial pose of the object $\bm{\xi}_0 \in SE(3)$ which can be obtained by any 6D pose estimation methods
    \item A sequence of RGB images $I_j,~j\in\{0,1,\cdots,N\}$
    \item A sequence of IMU data $G_k,~k\in\{0,1,\cdots,M\}$, where $G_k$ is a $6$ dimension vector consisting of the acceleration and angular velocity measurements along the three Cartesian axes of the IMU frame
\end{itemize}

In this section, we first provide a general overview of the pose tracking workflow that includes a pre-processing stage of raw image and IMU input prior to their use by the VIPose DNN. We describe afterward our novel VIPose DNN that infers the relative pose between consecutive frames by combining two consecutive images and a sequence of IMU data. The final object pose is then obtained by combining the relative transformations across consecutive frames. 

\subsection{Pose Tracking Workflow}
The pose workflow of our pipeline at time $t$ is illustrated in Fig.~\ref{fig:pipeline}. The camera provides the image observation $I_{t}$ and the IMU gives the acceleration and angular velocity. The image pre-processing is achieved in three main steps: rendering, cropping, and resizing and it is followed by an IMU pre-processing step. 

\subsubsection{Rendering}
First, we render an RGB image $I_{t-1}^R$ as the reference image of the last frame using the 3D model of the object and the estimated 6D pose of the last frame $\bm{\xi}_{t-1}$ to provide the target object to track. The rendering process extracts the object information from the cluttered scene. It is performed by placing the 3D model of the object at the center of a scene with a black background and projecting an RGB image at the viewpoint defined by the pose $\bm{\xi}_{t-1}$, as illustrated in Fig.~\ref{fig:bbox}. 

\subsubsection{Cropping}

\begin{figure}[t!]
    \centering
    \includegraphics[width=0.7\linewidth]{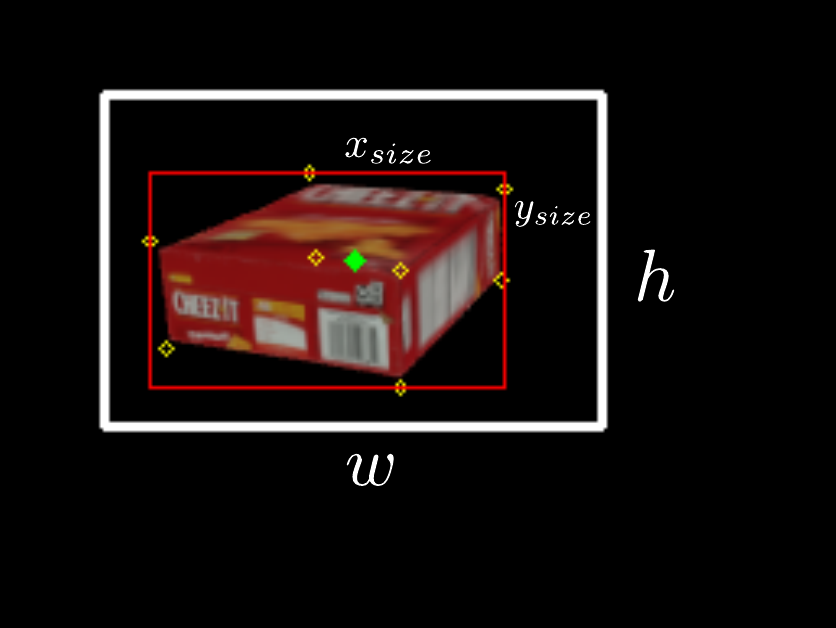}
    \caption{Visualization of Rendering and Cropping. The object is rendered using the 3D model of the object and estimated 6D pose. The 2D projection of the center and the 8 corners of the 3D model are illustrated as a green point and yellow points, respectively. The red box defines the size of the rendered object $x_{size}$ and $y_{size}$. The white box represents the enlarged bounding box which is centered at the green point with width $w$ and height $h$.}
    \label{fig:bbox}
        \vspace{-15pt}
\end{figure}

Due to the small object size in the raw image, it can be difficult to extract useful features for matching across frames. Inspired by~\cite{deepim}, we address this issue by using an enlarged bounding box (white box) as illustrated in Fig.~\ref{fig:bbox}  to crop both the rendered image $I_{t-1}^R$ and the image observation $I_{t}$. To obtain such bounding box, we first estimate the size of the object in the rendered image using the 2D projection of the $8$ corners of the corresponding 3D object model $p_i,~i \in {1,2,\cdots,8}$ as depicted with yellow points in Fig.~\ref{fig:bbox}
\begin{align*}
    x_{\text {size }} &= \max_i\left(p_{{i}_x}\right)-\min_i\left(p_{{i}_x}\right) \\
    y_{\text {size }} &=  \max_i\left(p_{{i}_y}\right)-\min_i\left(p_{{i}_y}\right)
\end{align*}
The enlarged bounding box is centered at the 2D projection of the center of the 3D model (depicted with the green point in Fig.~\ref{fig:bbox}) and keep the same aspect ratio of the input image. The height $h$ and width $w$ of the enlarged bounding box are obtained as
\begin{align*}
    h &= \lambda \cdot \max\left(x_{\text {size }} / r, y_{\text {size }}\right) \\
    w &= \lambda \cdot \max\left(x_{\text {size }}, y_{\text {size }} \cdot r\right)
\end{align*}
where $r$ is the aspect ratio of the original image and $\lambda$ denotes the expand ratio to ensure the box contains the object in both the rendered image and the new image.

\subsubsection{Resizing}
 Subsequently, we resize the cropped image pair to obtain more details. The cropped image pair is resized to a fixed resolution which is half of the original image size to reduce computation in the network. At the training stage, this enables a larger batch size for faster convergence of the training process. At the testing stage, this significantly reduces the running time of the system without compromising the accuracy. The resized image pair is used as the visual input to the network.
\begin{figure*}[t!]
    \centering
    \includegraphics[width=1.0\linewidth]{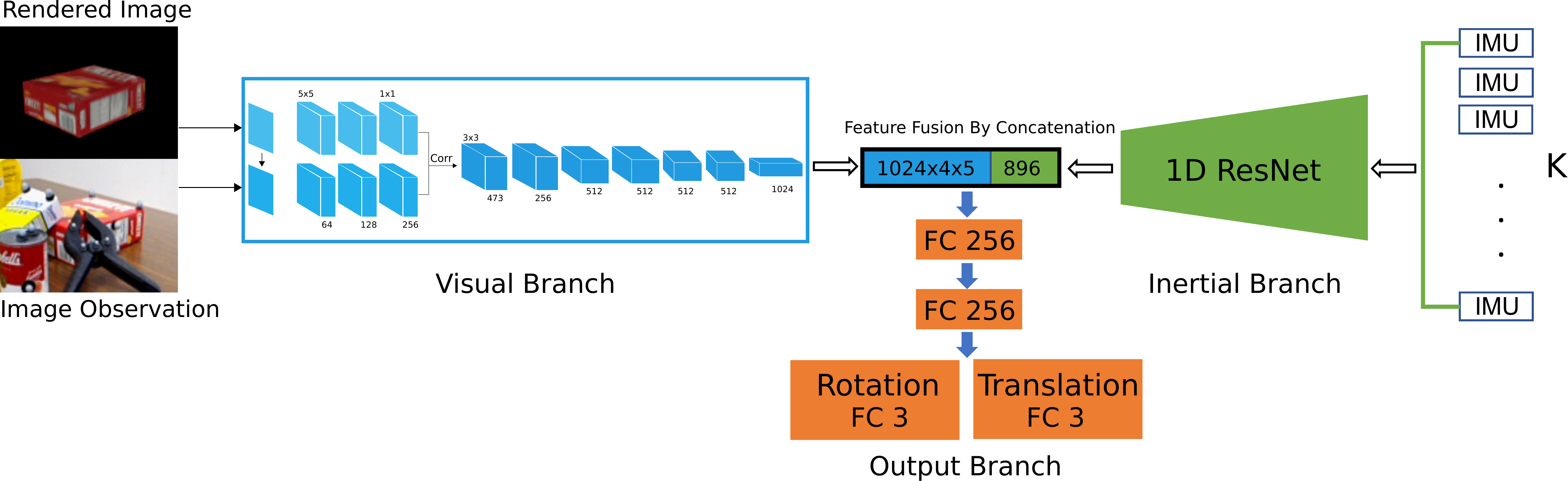}
    \caption{VIPose Network Architecture. The blue blocks represent the FlowNetC-based visual branch that produces the visual feature. The green block represents the 1D ResNet-based inertial branch that produces the inertial feature. The visual and inertial feature are fused by concatenating two feature vectors and provide a multi-modal feature for relative pose estimation. The orange blocks represent the output branch that directly regresses the relative pose between the two consecutive frames.}
    \label{fig:network}
        \vspace{-15pt}
\end{figure*}
 \subsubsection{IMU pre-processing}
After the image pair is pre-processed, we pre-process the IMU data as the inertial input to the network. An IMU buffer is designed as a queue that stores the most recent $K$ IMU measurements. The choice of $K$ can vary according to the frequency of the IMU input and the speed of the camera for different applications. When the new image observation arrives, the overall data in the IMU buffer is employed as input to the network and is represented by a $K \times 6$ tensor. The IMU sequence is rotated to a heading-agnostic coordinate frame from the IMU frame using ground-truth orientation directly from the device similar to~\cite{ronin}. The heading-agnostic coordinate frame is defined as a coordinate frame whose Z-axis is aligned with gravity and is consistent through the sequence.

The pre-processed image pair and IMU sequence are fed into the VIPose network to predict the relative object pose $\Delta\bm{\xi}_t\in SE(3)$ between the images $I_{t-1}^R$ and $I_{t}$. Then the 6D pose of the object in current image $\bm{\xi}_t$ is obtained as $\bm{\xi}_t=\Delta \bm{\xi}_t \otimes \bm{\xi}_{t-1}$, where $\otimes$ is matrix multiplication operation. The tracking process is repeated until the 6D pose of the last image is obtained.

In general, at time $t$, it is helpful to use the rendered image $I_{t-1}^R$ instead of the image observation $I_{t-1}$ as the reference image of the last frame for pose tracking. To further clarify this aspect, if the estimated object pose $\bm{\xi}_{t-1}$ is inaccurate, there is a mismatch between the object pose $\bm{\xi}_{t-1}$ and the corresponding object observation in the image $I_{t-1}$ since the pose $\bm{\xi}_{t-1}$ is affected by estimation drift. Conversely, the rendered image $I_{t-1}^R$ is a completely new image generated using the estimated object pose $\bm{\xi}_{t-1}$ and it does not suffer from the aforementioned observation issue. Therefore, estimating the relative pose between $I_{t-1}^R$ and $I_t$ is agnostic to the pose drift and contributes to reducing the global object pose estimation drift.

\subsection{VIPose Network}

The proposed network architecture is illustrated in Fig.~\ref{fig:network}. The network consists of a visual branch that takes as input the processed image pair $I_{t-1}^R$ and $I_{t}$, an inertial branch that extracts features from the IMU sequence, and an output branch that predicts the relative pose $\Delta \bm{\xi}_t$.

The visual branch (the blue block in Fig.~\ref{fig:network}), is based on FlowNet-C network~\cite{flownet} to extract visual features from the input image pair and we only adopt the convolutional layers of the original FlowNet-C network. The network is pre-trained to predict the optical flow between a pair of images. The representation of optical flow is considered useful for relative pose estimation as proved in \cite{deepim}. The first $3$ convolution layers extract low-level features from the two input images separately and produce two feature maps for the correlation layer that is responsible to compute the patch-wise similarity and therefore find the correspondence between the image pair. Then the last $6$ convolutional layers extract high-level features and reduce the size of the feature map. The feature map output of the last convolution layer is flattened to the visual feature vector for later feature fusion.

The inertial branch (the green block in Fig.~\ref{fig:network}) instead adopts the 1D version of the standard ResNet-18 network \cite{resnet} and takes as input the IMU sequence, which is the $K \times 6$ tensor from the IMU buffer as previously specified. Inspired by~\cite{ronin}, we employ a 1D ResNet which is pre-trained to predict the average 2D velocity given a sequence of IMU inputs. This has shown to have better performances with respect to other network architectures including LSTM~\cite{lstm} and Temporal Convolution Network~\cite{tcn}. The inertial branch initializes with pre-trained weights from~\cite{ronin}. Similarly, the output of the last 1D Convolution layer is flattened to the inertial feature vector. 

The visual feature and inertial feature are fused by concatenating the two feature vectors, which provide a multi-modal feature for regressing the relative transformation. The inertial feature provides useful information on camera motion for inferring the relative pose when the visual information of the target object is limited like for severe occlusions cases.

The output branch (the orange block in Fig.~\ref{fig:network}) is a multi-layer perceptron that takes as input the fused feature vector and outputs the relative pose between the two frames. It contains two fully-connected layers with $256$ dimensions, followed by two separate fully-connected layers for predicting the rotation and translation, respectively. The representation of the relative pose is crucial to the performance of the network. We adopt the se(3) representation in \cite{se3tracknet}, where $\mathbf{v}=\left[\mathbf{t}, \mathbf{w}\right]^\top \in se(3)$, such that its pseudo-exponential mapping lies in $SE(3)$
\begin{equation}
\Delta \bm{\bm{\xi}} = \text{pseudo-exp} (\mathbf{v})=\left[\begin{array}{ll}
\mathbf{R} & \mathbf{t} \\
\mathbf{0}^{\top} & 1
\end{array}\right] \in SE(3),
\end{equation}
where
$$
\mathbf{R}=\mathbf{I}_{3 \times 3}+\frac{\left[\mathbf{w}\right]_{\times}}{|\mathbf{w}|} \sin (|\mathbf{w}|)+\frac{\left[\mathbf{w}\right]_{\times}^{2}}{|\mathbf{w}|^{2}}\left(1-\cos \left(|\mathbf{w}|\right)\right),
$$
and $[\mathbf{w}]_{\times} $ is the skew-symmetric matrix. The last two output layers are used to directly regress $\mathbf{w}$ and $\mathbf{t}$ and the prediction is then used to compute the relative pose $\Delta \bm{\xi}$.
The network is trained end-to-end considering the following $L_2$ loss
\begin{equation}
\mathcal{L} = \lambda_{1}\cdot \left\|\mathbf{w}-\mathbf{w}_{gt}\right\|_{2}+\lambda_{2}\cdot \left\|\mathbf{t}-\mathbf{t}_{gt}\right\|_{2},
\end{equation}
where $\lambda_{1}, \lambda_{2}$ is the loss weight for rotation and translation, and $w_{gt}, t_{gt}$ denote the ground-truth relative pose.

\section{Experimental Results}~\label{sec:experiments}
In this section, we first introduce the VIYCB dataset and the automatic labeling technique. We describe afterward the training procedure and implementation details. We present the evaluation metrics and result on the VIYCB dataset lastly.

\subsection{Dataset}
\begin{figure}[t!]
    \centering
    \includegraphics[width=1.0\linewidth]{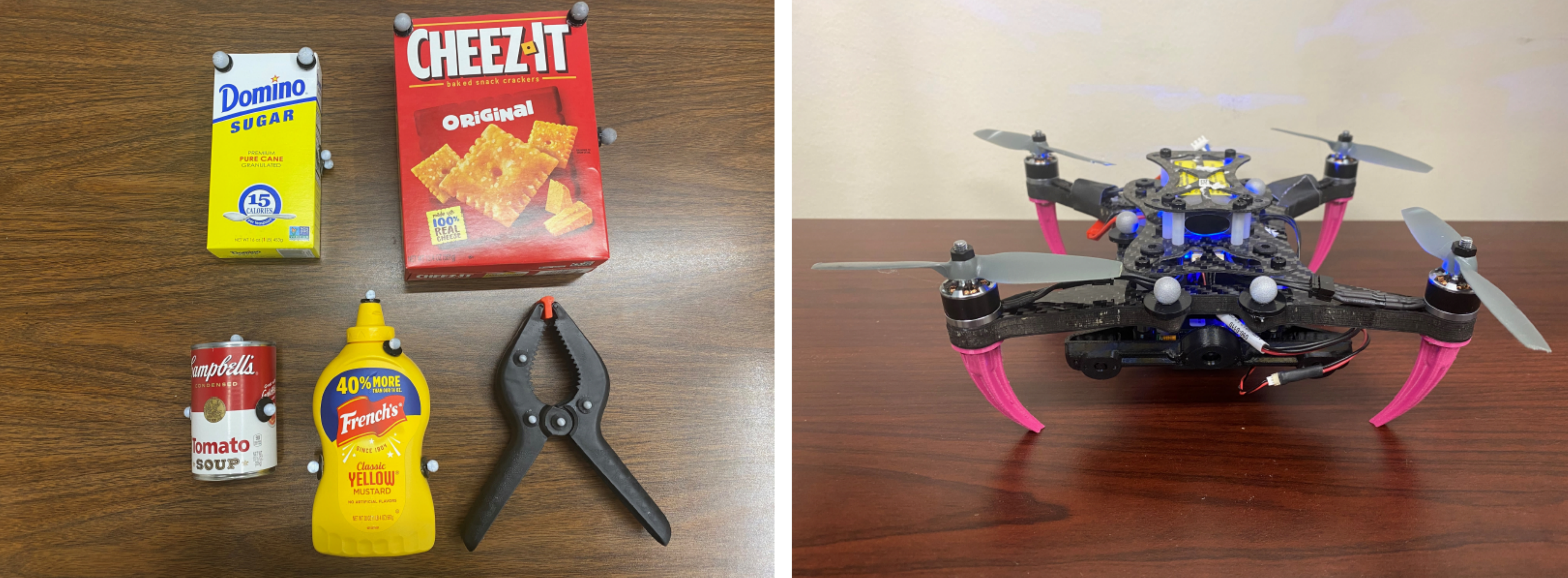}
    \caption{Left is the 5 YCB objects used in our dataset. Right is the quadrotor with the camera and IMU (hardware synchronized) for collecting the raw data. The grey markers on the objects and the quadrotor are used by the Vicon motion capture system for localization. }
    \label{fig:dataset_setup}
    \vspace{-20pt}
\end{figure}
To train and validate our method, we collect a video dataset named VIYCB dataset with RGB images, IMU data, and accurate 6D pose annotations. The dataset collection is conducted in the new indoor testbed $10\times5\times4~\si{m^3}$ at the ARPL lab at New York University. Fig.~\ref{fig:dataset_setup} shows the objects and the quadrotor equipped with a Qualcomm\textsuperscript{\textregistered} Snapdragon\textsuperscript{TM} board~\cite{LoiannoRAL2017} that we employed to collect the dataset. We use this device because the IMU and camera sensors are hardware synchronized. The objects in our experiment are $5$ of the YCB objects~\cite{ycbobject} which are widely accessible and cover different colors, shapes, scales, and object symmetries. We adopt an automatic labeling technique leveraging a Vicon\footnote{\url{www.vicon.com}} motion capture system to obtain the 6D pose annotation of all objects in each frame and save the effort of manually annotating the 6D pose of the objects. The Vicon system outputs the position and orientation of the objects and the camera in the Vicon world frame at $100~\si{Hz}$. The ground-truth 6D pose annotation is then obtained by transforming the 6D pose of the objects in the world frame to the camera frame. For each video, the 5 objects are placed closely in different arrangements which creates severe occlusions and even complete occlusions. The data is collected by moving around the objects with a handheld quadrotor. The RGB images ($480\times 640$ resolution) are recorded at around $25$ Hz, whereas the IMU sensor data is collected at $200$ Hz. The IMU orientation of the quadrotor is also recorded at $200$ Hz, similar to~\cite{ronin}. Generally, $8533$ images are used in training and $3142$ keyframes are used for evaluation.

\newcolumntype{g}{>{\columncolor{Gray}}c}
\begin{table*}[]
\centering
\begin{tabular}{|c|c|c|c|c|c|c|c|c|}
\hline
                       & \multicolumn{2}{c|}{PoseRBPF 50} & \multicolumn{2}{c|}{PoseRBPF 100} & \multicolumn{2}{c|}{DeepIM} & \multicolumn{2}{c|}{VIPose (Ours)} \\ \hline
Objects                & ADD             & ADD-S          & ADD             & ADD-S           & ADD          & ADD-S        & ADD                      & ADD-S                    \\ \hline
003\_cracker\_box      & 65.99           & 82.39          & 66.85           & 83.10           & \textbf{77.48}        & \textbf{87.62}        & 75.66                    & 86.84                    \\ \hline
004\_sugar\_box        & 62.29           & 81.04          & 63.19           & 82.05           & \textbf{78.92}        & \textbf{88.22}        & 74.46                    & 86.27                    \\ \hline
005\_tomato\_soup\_can & 55.10           & 78.64          & 56.36           & 80.04           & \textbf{76.22}        & \textbf{87.07}        & 71.54                    & 82.79                    \\ \hline
006\_mustard\_bottle   & 56.93           & 76.02          & 57.55           & 78.09           & \textbf{75.23}        & \textbf{85.89}        & 70.87                    & 82.12                    \\ \hline
051\_large\_clamp      & 44.42           & 66.18          & 45.22           & 67.60           & \textbf{62.37}        & \textbf{79.51}        & 59.68                    & 78.07                    \\ \hline
ALL                    & 56.95           & 76.85          & 57.83           & 78.20           & \textbf{74.04}        & \textbf{85.66}        & 70.44                    & 83.22                    \\ \hline
Speed (fps)            & \multicolumn{2}{c|}{21.9}        & \multicolumn{2}{c|}{12.3}         & \multicolumn{2}{c|}{8.7}    & \multicolumn{2}{c|}{\textbf{50.2}}   \\ \hline
Re-init (times)            & \multicolumn{2}{c|}{25}        & \multicolumn{2}{c|}{22}         & \multicolumn{2}{c|}{18}    & \multicolumn{2}{c|}{\textbf{9}}   \\ \hline
\end{tabular}
\caption{Result on VIYCB dataset. ADD and ADD-S are short for AUC (Area Under Curve) of ADD and ADD-S.}
\label{tab:result}
    \vspace{-10pt}
\end{table*}

\begin{figure*}[!t]
    \centering
    \includegraphics[width=0.8\linewidth]{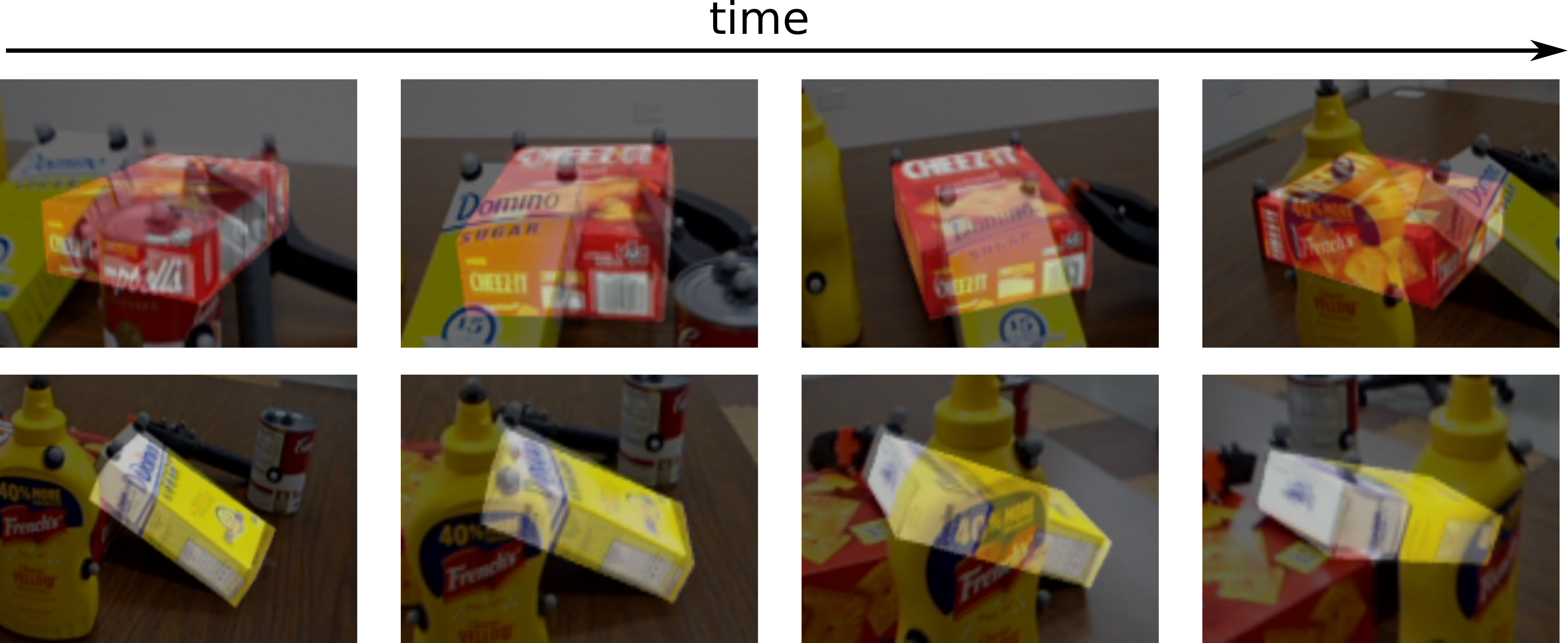}
    \caption{Qualitative Result of VIPose. The first row and the second row illustrate pose tracking of two objects, 003\_cracker\_box and 004\_sugar\_box, under severe occlusions. The target object is rendered in the scene using the estimated 6D pose of our method. This demonstrates the robustness of the proposed approach on handling the occlusion challenge.}
    ~\label{fig:qualitative_result}
        \vspace{-15pt}
\end{figure*}

\subsection{Architecture Implementation Details}
For network training, we adopt a two-step training strategy. We first train the visual branch only. Subsequently, we jointly train the visual, the inertial, and the output branches.

The visual branch takes a pre-processed image pair as input, which contains a real image $I_t$ and a rendered image $I_{t-1}^R$. The training data consists of image pairs from collected data and synthetic data. For an image $I_t$ and corresponding ground-truth pose $\bm{\xi}_t$ in collected data, an image pair is generated by rendering a image $I_{t-1}^R$ with the ground-truth pose of the last frame $\bm{\xi}_{t-1}$, cropping both $I_{t-1}^R$ and $I_t$ with the enlarged bounding box introduced in Section~\ref{sec:approach}, and resizing the cropped image pair to $240\times 320$. In addition, we can generate synthetic image pairs to build a more diverse training dataset that covers a wider range of relative pose distributions. Instead of using ground-truth pose of last frame $\bm{\xi}_{t-1}$, we randomly sample a relative pose $\Delta \bm{\xi}_t^{*}$ and compute the object pose in the last frame $\bm{\xi}_{t-1}^{*}$ which is used to render $I_{t-1}^{R*}$. The image pair $I_{t-1}^{R*}$ and $I_t$ are then cropped and resized as described above. The sampled relative pose $\Delta \bm{\xi}_t^{*}$ consists of a rotation part and a translation part. The rotation is obtained considering three Euler angles sampled from a Gaussian distribution $\mathcal{N}\sim\left(0,10^{2}\right)$. For the translation part, each of the three components is sampled as well from a Gaussian distribution $\mathcal{N}\sim\left(0,0.02^{2}\right)$.
We only train the visual branch and we leverage a pseudo output branch to train the visual branch to adapt the change in the the dimension of the first fully-connected layer of the output branch. Each training batch consists of image pairs from both collected data ($1/8$) and synthetic data ($7/8$) and the batch size is set to 64. The visual branch and the pseudo output branch are trained using an SGD optimizer for $30$ epochs with an initial learning rate of $0.001$. A cosine annealing scheduler reduces the learning rate after each epoch, to $0.0001$ at the end.

Subsequently, we jointly train the visual branch, the inertial branch, and the output branch. The training data in this step is only the image pairs from collected data with corresponding IMU sequence due to the absence of IMU sequence in the synthetic data. The length of the IMU buffer $K$ is set to $200$ for both training and testing, which means it contains all IMU data within the last second. The ground-truth orientation is used to rotate the IMU data to the heading-agnostic coordinate frame from the IMU frame in both training and testing following \cite{ronin}. The pseudo output branch is replaced with a new output branch with randomly initialized weights and bias. The network is trained using an SGD optimizer for another $20$ epochs with an initial learning rate of $0.0001$ for both the visual and the inertial branch and $0.001$ for the output branch. Similarly, a cosine annealing scheduler reduces the learning rate after each epoch, to $0.00001$ and $0.0001$ at the end, respectively.

To avoid the over-fitting issue during training, we adopt data augmentation strategies including random HSV shift, Gaussian noise, and Gaussian blur on the real image $I_t$ in each image pair in both training steps. The expand ratio $\lambda$ is fixed to $1.4$ in the cropping step. The loss weight $\lambda_1$ and $\lambda_2$ is always set to $1$. A seperate model is trained for each object. The rendering process is implemented in OpenGL. All network training and testing are performed on an NVIDIA RTX 2080 Ti GPU.

\subsection{Evaluation}
\vspace{-2pt}
For pose estimation evaluation, We adopt the ADD \cite{linemod} and ADD-S \cite{posecnn} metrics 

\begin{align*}
\text{ADD} &=\frac{1}{m} \sum_{\mathbf{x} \in \mathcal{M}}\left\|(\mathbf{R} \mathbf{x}+\mathbf{T})-(\tilde{\mathbf{R}} \mathbf{x}+\tilde{\mathbf{T}})\right\|, \\
\text{ADD-S} &=\frac{1}{m} \sum_{\mathbf{x}_{1} \in \mathcal{M}} \min _{\mathbf{x}_{2} \in \mathcal{M}}\left\|\left(\mathbf{R} \mathbf{x}_{1}+\mathbf{T}\right)-\left(\tilde{\mathbf{R}} \mathbf{x}_{2}+\tilde{\mathbf{T}}\right)\right\|,
\end{align*}
where $\mathcal{M}$ represents the points of the 3D object model, $m$ denotes the number of the points $\mathcal{M}$, $(\mathbf{R}, \mathbf{T})$ and $(\tilde{\mathbf{R}}, \tilde{\mathbf{T}})$ denote the rotation and translation of ground-truth pose and estimated pose, respectively. The ADD metric computes the average distance between the transformed 3D points using ground-truth pose and estimated pose, respectively. The ADD-S metric is specifically designed for symmetric objects where the average distance is computed by closest point distance. The estimated pose is considered correct when the result of ADD or ADD-S is smaller than a given threshold. We compute the AUC (Area Under Curve) of ADD and ADD-S results where the threshold varies for the average distance and the maximum threshold is set to $10$ cm.
\subsection{Results on VIYCB Dataset}
\vspace{-2pt}
Table \ref{tab:result} shows the pose tracking results on the VIYCB dataset. The proposed approach is compared with other state-of-the-art RGB-based object pose tracking methods including DeepIM \cite{deepim} and PoseRBPF \cite{poserbpf}. All compared methods use the ground-truth pose for initialization at start. VIPose and DeepIM are re-initialized with ground-truth pose when the average relative pose prediction in the last $10$ frames is larger than a threshold, specifically its rotation greater than $10$ degrees or its translation greater than $1$ cm. PoseRBPF with $50$ particles and $100$ particles are evaluated for comparison. PoseRBPF utilizes the ground-truth pose to initialize its detection center and motion prior. It is re-initialized when the target object is heavily occluded, which is detected when the maximum similarity between the feature embedding and the codebooks among all particles is lower than a pre-defined threshold ($0.6$ in the presented case). Since the objects in our dataset are the subset of the YCB video dataset, we use the pretrained models of DeepIM and PoseRBPF to perform pose tracking on our VIYCB dataset for evaluation. 

In the experiment, for each object, we run each method to track the 6D pose of the object in all testing videos and we evaluate the aforementioned metrics on the keyframes. We also compare the speed of these methods using their average computation time for each frame. To evaluate the robustness of these methods, we compare the frequency of re-initialization operations in all testing videos and average it for each object class since most re-initialization operations are caused by heavy occlusions.

Results in Table \ref{tab:result} show that the proposed method performs consistently better than PoseRBPF with $50$ particles and $100$ particles on all $5$ objects in terms of accuracy and speed. The accuracy of VIPose is comparable to the DeepIM approach, while our approach is substantially faster and runs at $50.2$ Hz compared to $8.7$ Hz of DeepIM. This indicates that VIPose is more suitable for real-time applications. The speed-accuracy trade-off is presented in Fig.~\ref{fig:fig1}. 

In terms of robustness to occlusions, VIPose only requires re-initialization every $349$ frames when the target object is completely occluded. PoseRBPF with $50$ and $100$ particles requires re-initialization more frequently for every $126$ frames and $143$ frames, whereas DeepIM on average every $174$ frames. This indicates that the proposed method is more robust to occlusion than previous methods and proves the effectiveness of the inertial feature in the pose tracking task.
\vspace{-10pt}
\section{Conclusion}~\label{sec:conclusion}
In this work, we proposed the first real-time RGB-based 6D pose tracking approach called VIPose DNN that combines inertial and camera information The results show that our method has comparable performance with respect to existing approaches while being real-time and robust to heavy occlusions. Furthermore, it provides remarkable pose estimation results for heavily occluded objects.

Future works will consider how to extend this network to the multi-objects pose tracking task. Moreover, we would also like to investigate the use of adaptive feature fusion to better combine inertial and camera information in different scenarios to increase pose tracking accuracy and robustness to different testing scenarios without sacrificing the real-time capabilities of our approach.

\bibliographystyle{IEEEtran}
\bibliography{mybib}
 \end{document}